\documentclass[runningheads]{llncs}
\usepackage{graphicx}
\usepackage{amsmath,amssymb} 
\usepackage{color}

\begin{document}

\newcommand{\point}{
    \raise0.7ex\hbox{.}
    }


\pagestyle{headings}

\mainmatter

\title{A Holistic Approach for Data-Driven Object Cutout} 

\titlerunning{A Holistic Approach for Data-Driven Object Cutout} 

\authorrunning{Huayong Xu et al.} 

\author{
	Huayong Xu\inst{1} \and
	Yangyan Li\inst{3} \and
	Wenzheng Chen\inst{1} \and
    	Dani Lischinski\inst{2} \and
	Daniel Cohen-Or\inst{3} \and
	Baoquan Chen\inst{1}
} 

\institute{
	Shandong University \and
	Hebrew University of Jerusalem \and
	Tel Aviv University
} 

\maketitle

\begin{abstract}
Object cutout is a fundamental operation for image editing and manipulation, yet it is extremely challenging to automate it in real-world images, which typically contain considerable background clutter. In contrast to existing cutout methods, which are based mainly on low-level image analysis, we propose a more \emph{holistic} approach, which considers the entire shape of the object of interest by leveraging higher-level image analysis and learnt global shape priors. Specifically, we leverage a deep neural network (DNN) trained for objects of a particular class (chairs) for realizing this mechanism. Given a rectangular image region, the DNN outputs a probability map (P-map) that indicates for each pixel inside the rectangle how likely it is to be contained inside an object from the class of interest. We show that the resulting P-maps may be used to evaluate how likely a rectangle proposal is to contain an instance of the class, and further process good proposals to produce an accurate object cutout mask. This amounts to an automatic end-to-end pipeline for catergory-specific object cutout. We evaluate our approach on segmentation benchmark datasets, and show that it significantly outperforms the state-of-the-art on them.

\end{abstract}

\section{Introduction}
\label{sec:introduction}

Object cutout is a fundamental operation in image editing and manipulation~\cite{Chen_SIGGRAPHAsia09_Sketch2Photo,Xu_SIGGRAPH11_Photo,Zheng_SIGGRAPH12_Interactive}, an operation which graphics artists perform routinely. Performing this operation in a completely automatic fashion involves solving two classical and challenging computer vision tasks: object detection and semantic segmentation. Furthermore, in some scenarios an automatic approach is infeasible, since the user's intent is difficult to predict. Thus, a variety of interactive cutout tools have been proposed over the years, e.g.,~\cite{Mortensen:1995,Li_SIGGRAPH04_Lazy,Rother_SIGGRAPH04_GrabCut}. A common approach is to let the user indicate the object of interest with a bounding box, and attempt to proceed automatically from this minimal input to obtain an accurate cutout mask~\cite{Rother_SIGGRAPH04_GrabCut}.

\begin{figure}[t!]
\centering
\includegraphics[width=\linewidth]{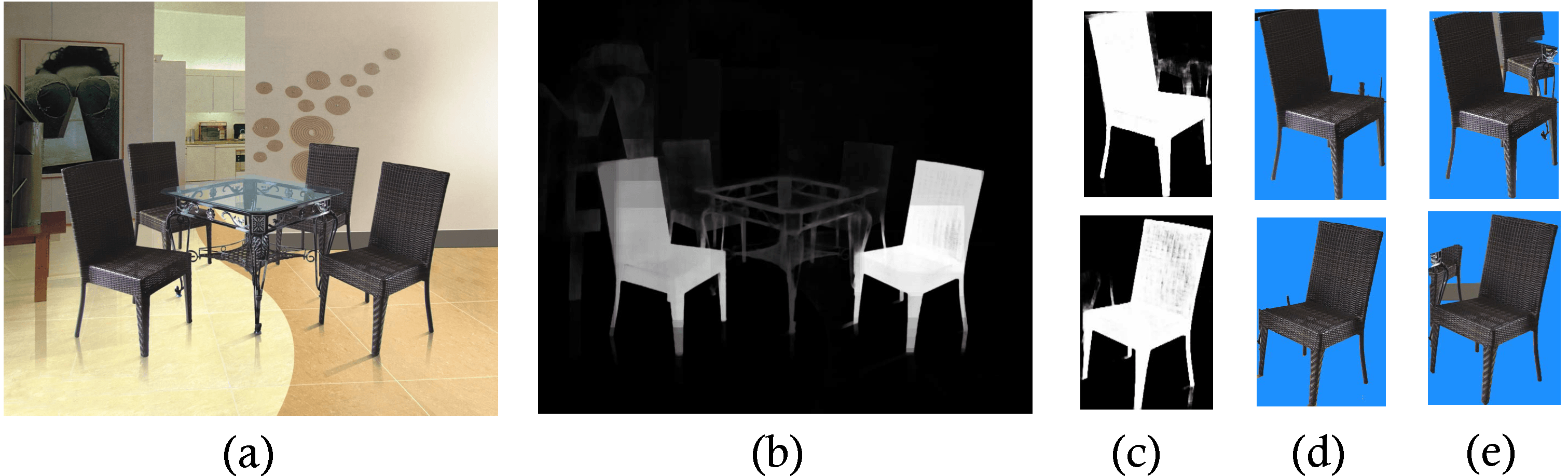}
\caption{
A cluttered scene with four chairs (a); an aggregated P-map visualizing object detection (b); local P-maps inside proposed rectangles (c) ; cutouts produced with the aid of our local P-maps (d) ; cutouts produced using GrabCut, for the same rectangles (e) .
}
\label{fig:teaser}
\end{figure}

However, these tasks of detection and segmentation, which the human visual system accomplishes with ease, are notorious for being surprisingly hard for a computer program. They are especially challenging when the object of interest is located in front of a cluttered background, which may contain many distractions, such as other objects with similar low-level statistics to the foreground object. Such an example is demonstrated in Fig.~\ref{fig:teaser}(a), where the background contains chairs identical in appearance to those in the foreground.

Methods that are based mainly on low-level image analysis tend to fail when the foreground object and the background are not statistically separable, and when salient separating edges cannot be easily detected. Sparse user input, such as a bounding box \cite{Rother_SIGGRAPH04_GrabCut} or a pair of scribbles \cite{Li_SIGGRAPH04_Lazy}, is not sufficient to overcome these difficulties, as demonstrated in Fig.~\ref{fig:teaser}(e). A more \emph{holistic} approach, which considers the whole shape rather than its pieces by leveraging higher-level image analysis and global object shape priors, has a better chance of coping with these challenging scenarios.

Recent advances in deep neural networks (DNNs) have shown promising results in solving various image understanding tasks, such as classification, detection and segmentation~\cite{Russakovsky_IJCV15_ImageNet}. However, object cutout presents DNNs with three additional challenges. Firstly, the network should learn a large variety of detailed shape priors, which differ significantly among different object classes. Secondly, the solution space is high-dimensional, since the images operated upon and the resulting cutout masks are required to be of high resolution. Thirdly, the cutout masks have sharp boundaries. For example, the state-of-the-art DNN-based instance-level object segmentation approach of~\cite{Liang_arXiv15_Reversible} achieves $24.5\%$ $AP^r$ at $0.5$ $IoU$\footnote{$AP$ is short for \emph{average precision}, which is the area under precision-recall (PR) curve. $IoU$ is short for Intersection over Union, i.e., ${A(P \bigcap G)}/{A(P \bigcup G)}$, where $P$ and $G$ are segmentation prediction and ground truth, respectively, while $A(\bullet)$ indicates their areas. To measure the precision of segmentation, $AP^r$ is used, which is \emph{region} based $AP$. Here, a segmentation is considered to be positive when it reaches $0.5$ $IoU$.}, on the chair class, which is far from being useful for graphics applications.

In this work, we leverage a Convolution-Deconvolution (DeconvNet) DNN~\cite{Noh_ICCV15_Learning}. However, we train it specifically using objects of a particular class (chairs). By focusing the training on a particular class, we reduce the learning difficulty and push it to learn more detailed shape priors. Moreover, to provide a more exhaustive coverage of the class in the training phase, we leverage synthetic imagery generated from ShapeNet~\cite{shapenet}. Given a rectangular image region, the trained network generates a map (of the same resolution as the input region), where each pixel indicates the likelihood of belonging to the object. We refer to such maps as \emph{P-maps} for short.

We show that the resulting P-maps are useful for a number of vision tasks and applications.

First, given a set of proposals (generated by any state-of-the-art method), we are able to evaluate and rank it better using the P-map. This capability enhances automatic location of chairs in an image. Second, getting back to the original motivation for our work, we are able to use the P-map to guide an iterative graphcut process \cite{Rother_SIGGRAPH04_GrabCut} towards an accurate object cutout (see Fig. \ref{fig:teaser}(d)). Thus, the approach described in this paper amounts to an end-to-end solution for automatic object cutout.

We use chairs as our running example, as they represent a family of shapes that have a rich variability of geometry and topology, and pose a challenge to state-of-the-art DNNs. Our technique is specifically designed to deal with cluttered images, learning to extract the foreground shape from a background that may contain objects with similar local statistics. We show that our holistic shape prior based approach considerably improves the accuracy of the resulting cutouts, compared to the current state-of-the-art, especially for cluttered images.

Do not use any additional Latex macros. All the individual papers need to be
merged into one volume, what requires that there are no conflicting Latex definitions.
Just plain ``basic Latex'', please.

\section{Related Work}
\label{sec:related_work}

Over the past few decades, tremendous amount of research have been devoted to studying how to faithfully perceive objects in images. Significant progress has been made on several sub-tasks towards this goal, including object recognition, object detection, and semantic segmentation, from which still only a coarse understanding of the scene can be established. In this section, we briefly review advances made in these directions and discuss their connections to the task of instance-level object cutout.

\paragraph{Image segmentation} is the process of partitioning an image into multiple segments of similar appearance. The problem can be formulated as a clustering problem in color space~\cite{Comaniciu_PAMI02_Mean}. To incorporate more spatial constrains into the process, the image may be modeled as a graph, converting image segmentation into a graph partition problem. The weights on the graph edges can either be inferred from pixel colors~\cite{Felzenszwalb_IJCV04_Efficient} or from sparse user input, as an addition~\cite{Rother_SIGGRAPH04_GrabCut}. Algorithms have been proposed for efficiently computing the partition, even when the pixels are densely connected (DenseCRF)~\cite{Krahenbuhl_NIPS11_Efficient}. Such methods are capable of inferring a sharp segmentation mask from sparse of fuzzy probabilities, and thus are widely used as a post-process for methods that produce segmentation probability maps.

\paragraph{Semantic segmentation.} Instead of grouping pixels only by appearance, semantic segmentation forms segments by grouping pixels belonging to same semantic objects; thus, a single segment might contain heterogeneous appearances. Since such segmentation depends on semantic understanding of the image content, state-of-the-art methods operate by running classification neural networks on patches densely sampled from the image in order to predict the semantic label of their central pixels~\cite{Long_CVPR15_Fully,Papandreou_ICCV15_Weakly,Zheng_ICCV15_Conditional}. Instead, Noh et al.~\cite{Noh_ICCV15_Learning} proposed a DeconvNet to directly output a high resolution semantic segmentation. We leverage DeconvNet for solving the more challenging object cutout problem by adapting and training it extensively on objects from a specific class.

\paragraph{Object cutout.} Object cutout further pushes semantic segmentation from category-level to instance-level. The additional challenge is that objects with similar appearance may hinder the cutout accuracy for individual instances. The state-of-the-art addresses the object cutout problem by solving it jointly with detection~\cite{Hariharan_ECCV14_Simultaneous,Liang_arXiv15_Reversible}, object number prediction~\cite{Liang_arXiv15_Proposal}, or by explicitly modeling the occlusion interactions between different instances~\cite{Silberman_ECCV14_Instance,Chen_CVPR15_Multi}. Though significant progress has been made recently, the performance on some object categories is still very low. In this work, we take advantage of being able to utilize training data synthesized from 3D models~\cite{Su_ICCV15_Render}, and focus on leveraging rich holistic shape priors for addressing segmentation ambiguities.

\paragraph{3D object retrieval and view estimation.} Recently, exciting advances in image based 3D object retrieval and object view estimation have made~\cite{Aubry_CVPR14_Seeing,Li_SIGGRAPHAsia15_Joint,Su_ICCV15_Render}. Such efforts are quite related to object cutout, as the retrieved 3D model can be rendered in the estimated view to approximate the object in the image, thus providing a strong prior for cutout. However, we found that the gap between projected proxies and accurate cutout masks cannot be easily bridged. One reason is that there are only few models in the existing shape databases that match well with real world objects. The inherent mismatch between 3D database and real world objects, plus the introduced retrieval and view estimation errors, render it infeasible to compute object cutout through such an approach, in general cases.

\paragraph{Object detection.} Object detection is usually done in two steps: object bounding box proposal generation and proposal evaluation. Proposal generation yields a set of bounding boxes that potentially contain objects~\cite{Uijlings_IJCV13_Selective,Zitnick_ECCV14_Edge,Krahenbuhl_CVPR15_Learning}. Proposal evaluation typically extracts features from the image patches contained in the proposed bounding rectangle, and estimates the confidence of the image patches to belong to objects of certain classes. R-CNN~\cite{Girshick_CVPR14_Rich} is an representative work in object detection and several works extended it to further improve efficiency and accuracy~\cite{He_ECCV14_Spatial,Girshick_ICCV15_Fast,Ren_NIPS15_Faster}. We show that the P-maps generated by our category-specific DeconvNet can benefit proposal evaluation for improving object detection and subsequent object cutout.


\section{Instance Probability Maps}
\label{sec:instance_cutout_dnn}

In this section, we introduce our method for generating instance probability maps. The term ``instance'' indicates that the maps aim to locate specific instances of a particular object class, rather than only detect the presence of such an object in the image. These probability maps, which will be referred to as P-maps, specify for each pixel its likelihood of belonging to an object instance. As we show in later sections, they allow efficient detection and consequent cutouts of objects, as well as the retrieval of 3D shapes.

Our P-maps are based on the non-trivial observation that although an image of an object may be high-dimensional, the underlying object can often be represented by a compact feature vector. Dosovitskiy et al.~\cite{Dosovitskiy_CVPR15_Learning} show that a DNN can be trained to generate object images from given object type, viewpoint, and color. This raises the expectation that neural networks can detect the presence of an object, encode it into a rather low-dimensional feature vector, from which it then should be possible to ``reconstruct'' the object, or its binary cutout mask. The premise of this approach is that the extraction of this low-dimensional representation in fact ``peels off'' the background clutter.

\begin{figure}[t!]
\centering
\includegraphics[width=0.8\columnwidth]{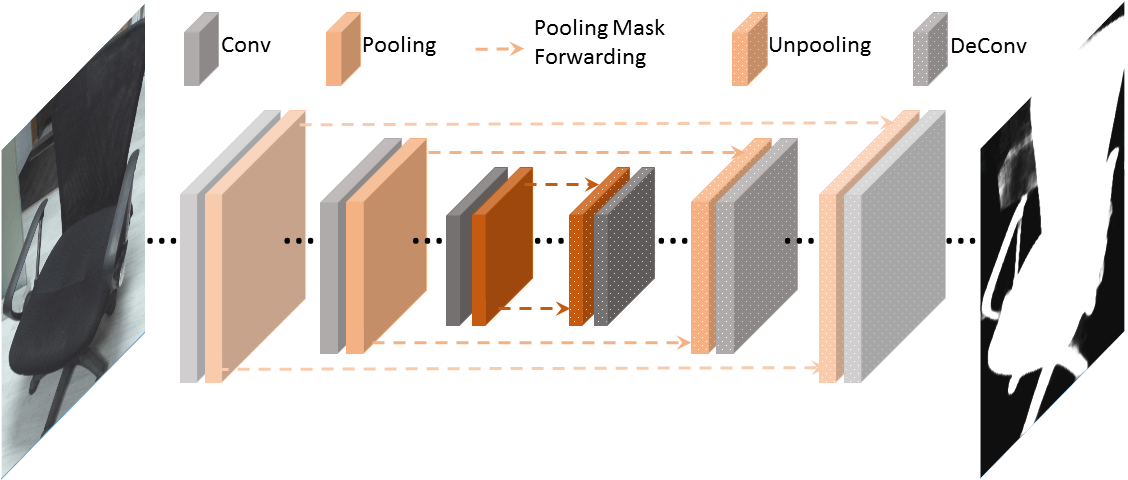}
\caption{A schematic illustration of the DeconvNet architecture. Records of the max pooling operations that occur during the first convolutional half, are forwarded to the subsequent deconvolution half of the network.}
\label{fig:conv}
\end{figure}

However, only rather fuzzy images can be reconstructed if the feature vector is extracted from real-world cluttered images, instead of a clean feature vector consisting of object type, viewpoint, and color~\cite{Dosovitskiy_arXiv15_Inverting}.
To generate a sharper image or cutout mask, additional information must be passed into the reconstruction process, and we build our approach upon the DeconvNet architecture proposed by Noh et al.~\cite{Noh_ICCV15_Learning}, which we found to be better suited for cluttered scenes. In this network, not only the feature vector, but also additional information about the feature extraction process is forwarded into the reconstruction process, which greatly improves the reconstruction sharpness.

\begin{figure}[t!]
\centering
\includegraphics[width=0.8\columnwidth]{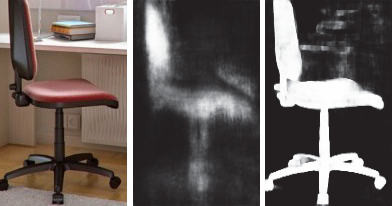}
\caption{. Given an input image (left), DNN trained extensively with large amount of images from a particular class can learn to ``reconstruct'' a fuzzy image while ignoring background clutters (middle). Pooling mask forwarding in DeconvNet greatly improves the sharpness of output cutout probability maps (right). }
\label{fig:pooling_mask}
\end{figure}

More specifically, the feature extraction part of our network (see Fig. \ref{fig:conv}) is composed of convolutional layers and pooling layers, which gradually encode the input as a 4096-dimensional feature vector. This feature vector is then taken by the reconstruction part of the network composed of deconvolutional layers and unpooling layers, which gradually reconstruct the P-map. Importantly, the pooling masks, which record the full history of the pooling operations, are forwarded into the unpooling layers. The pooling mask forwarding relieves the difficulty in learning how to perform a sharp reconstruction, thus greatly outperforming approaches that only use the feature vector. See Fig.~\ref{fig:pooling_mask} for a visual comparison of results with and without the use of pooling masks.

\begin{figure*}[t!]
\centering
\includegraphics[width=\linewidth]{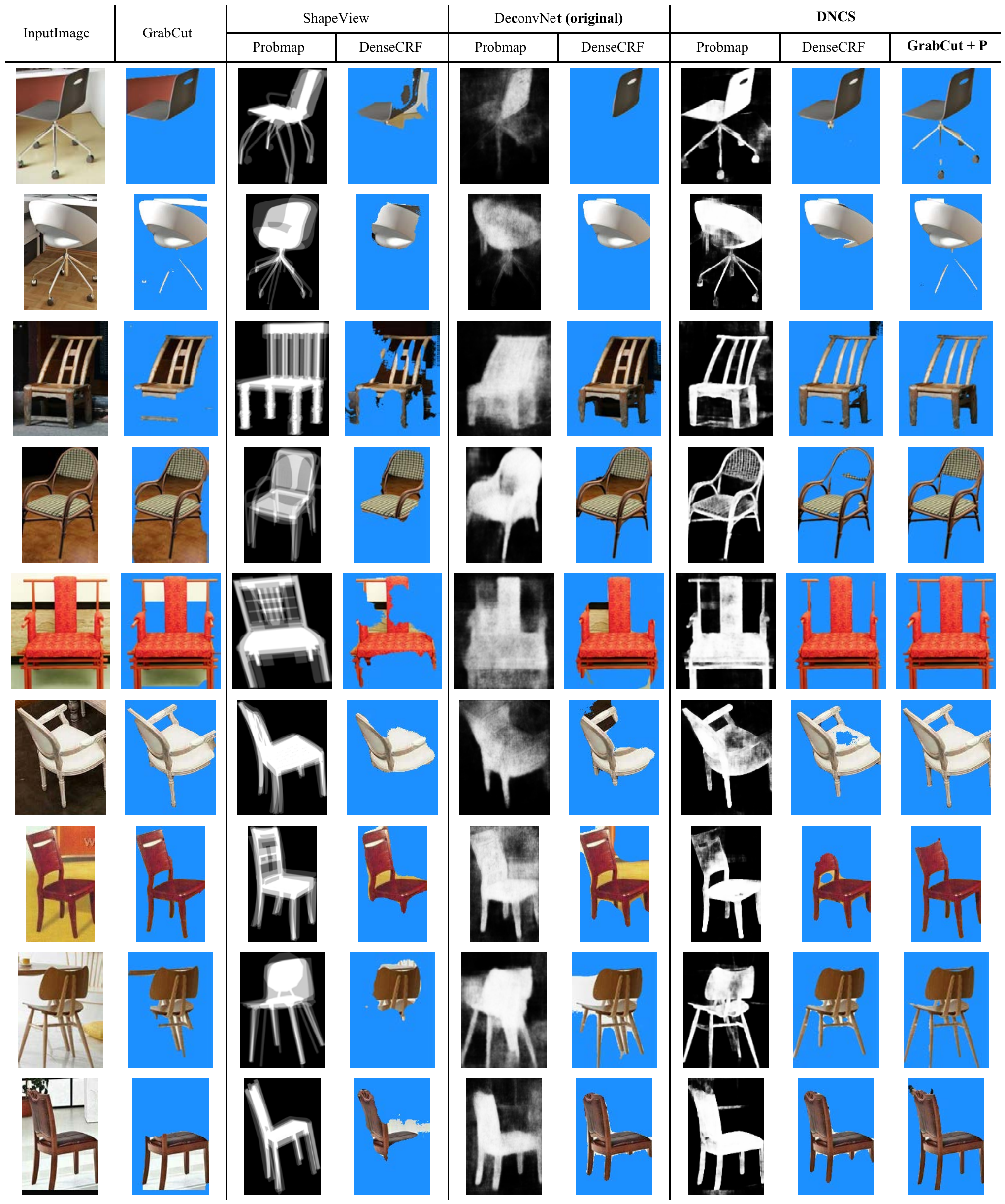}
\caption{Comparison of instance probability maps and the resulting cutout masks generated by various baseline methods and by our approach. It may be seen that our DNCS is more successful at injecting the learnt shape priors into the probability map generation. Furthermore, our GrabCut+P cutout method makes more effective use of the probability maps to produce a cutout, compared to DenseCRF.}
\label{fig:comparison}
\end{figure*}

The original DeconvNet was proposed for solving a semantic segmentation problem using $21$ classes. We adapt it to solve our instance-level segmentation problem by changing its last layer to output only two channel images: one for foreground and one for background. Then a softmax function over these two channels gives the foreground/background probability for each pixel. DeconvNet was originally trained on PASCAL VOC 2012~\cite{Everingham_IJCV09_The} data, where the number of segmented images is not particularly high, since image segmentation is a hard task for crowd sourcing. When narrowing the data to a specific category, it is insufficient to inject enough shape priors into the trained model. Instead, we choose to train the network using a much larger number of synthetic images with ground truth cutout masks, which are generated completely automatically by rendering 3D models. In the reminder of the paper, we refer our adapted DeconvNet as DNCS (DeconvNet-Class-Specific). As we shall see, the amount and quality of our training data enables the trained network to learn a powerful shape prior, which makes it possible to perform well even in the presence of considerable background clutter.

\section{Proposal Evaluation}
\label{sec:object_detection}

The ability to generate high-quality instance probability maps over rectangles of roughly the expected object size in the image is useful not only for generating accurate binary cutout masks (Section \ref{sec:grabcut}), but also helpful for locating object instances from a given scene image, referred to as \emph{detection task} in computer vision. Given a proposal, we are able to evaluate and rank it better when using the corresponding P-map, thus improve detecting object out of an entire image.

\begin{figure*}[t!]
\centering
\includegraphics[width=\linewidth]{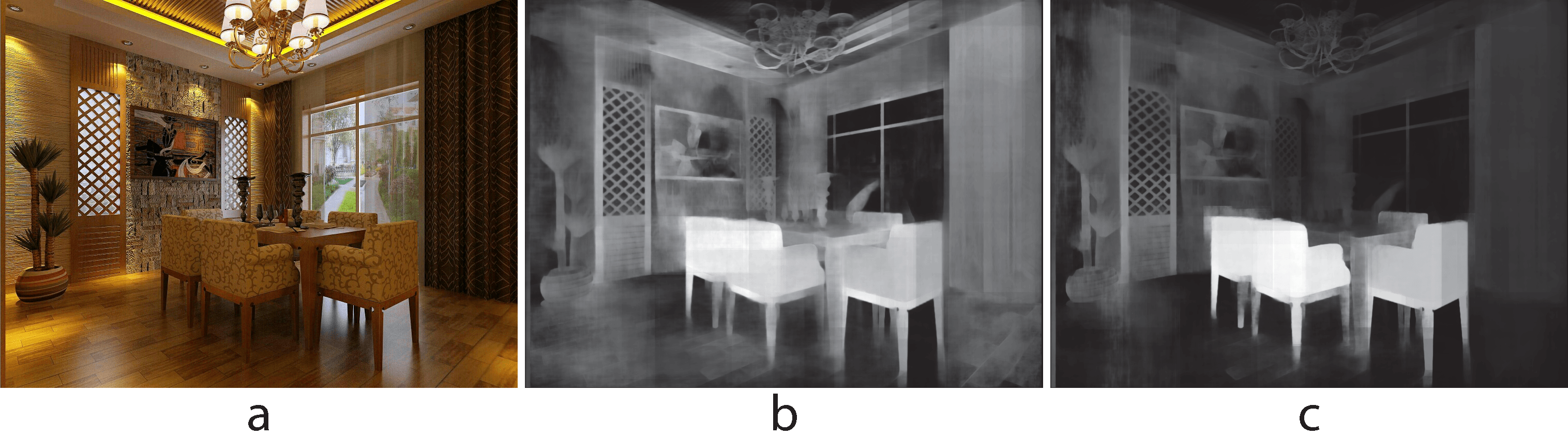}
\caption{An aggregated P-map for an entire image can be generated by accumulating instance probability map from bounding box proposals (b). By weighting the proposals with $\mathcal{X}_\textit{CNN}$ an even better aggregated P-map can be generated (c).}
\label{fig:aggregated_probmap}
\end{figure*}

\begin{figure*}[t!]
	\centering
	\includegraphics[width=\linewidth]{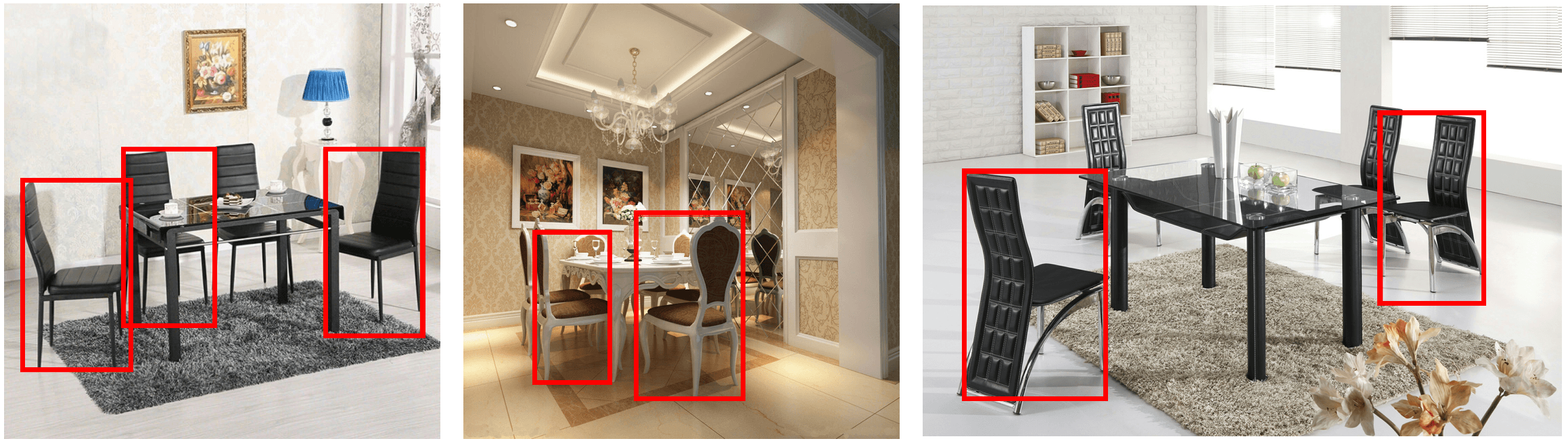}
	\caption{P-map enhanced chair detection results. Note that since our P-map ``sees'' the individual chairs, it can locate chairs well, even with heavy background clutter.}
	\label{fig:proposals}
\end{figure*}

\paragraph{Proposal evaluation on RGB-P images.} A proposal is a rectangular region in a large image, which is deemed likely to contain an object of interest. There are many methods that generate proposals, whose objective is to avoid performing an exhaustive search over the entire image. We show that using an RGB-P image, where the fourth P channel is computed by the instance cutout DNN, benefits such proposal evaluation methods. More formally, let $I_b$ be a rectangular proposal, its evaluation by a function $\mathcal{X}: I_b \rightarrow \mathcal{R}$, maps the input proposal to a real value that indicates the confidence of having an object of a specific class contained in it. In our case, the function $\mathcal{X}$ is no more than a binary classifier that tells how likely the proposal depicts a chair.

We train the classifier $\mathcal{X}$ with synthetic images, and we generate many rectangular proposals with any state-of-the-art methods. Since in our synthetic images we know the ground truth bounding boxes of the objects, we can easily generate positive and negative examples. We treat proposals with more than $80\%$ overlap with the ground truth bounding boxes as positive samples, and the rest as negative samples. For each proposal, we also compute its P-channel.

We trained two classifiers: $\mathcal{X}_\textit{SVM}$ and $\mathcal{X}_\textit{CNN}$. For $\mathcal{X}_\textit{SVM}$, we extract AlexNet~\cite{Krizhevsky_NIPS12_Imagenet} CNN features ($4096$ dimensions, the output of $fc7$ layer) from the RGB channels, and HoG~\cite{Dalal_CVPR05_Histogram} features ($24304$ dimensions) from the P-map, which are then reduced with PCA to $4096$ dimensions. Then we concatenate the CNN features and the PCA reduced HoG features for training a linear Support Vector Machine (SVM).
The $\mathcal{X}_\textit{CNN}$ classifier is an end-to-end CNN approach, where we add a fourth channel to the filters of the first convolutional layer of AlexNet to adapt the additonal P-channel, and fine tune the network to work as a binary classifier.

The effect of proposal evaluation is visualized in an aggregated P-map in Fig.~\ref{fig:aggregated_probmap}, where we generate an aggregated P-map, by running instance cutout in all proposals, accumulating the resulting P-maps with weights from the confidences given by $\mathcal{X}_\textit{CNN}$, and normalizing the result. Another example of such a map is shown in Fig. \ref{fig:teaser}(b). It is clear that our P-map enhanced proposal evaluation can greatly narrow down attentions to chair regions. We compare the performance of our two classifiers with versions trained without using the P channel, and found that both classifiers perform better when P channels is used (see Table~\ref{tab:proposal_evalution_comparison}). This is a strong evidence that the P-channels are effectively improving the proposal evaluation. As can be seen from Fig.~\ref{fig:proposals}, chairs, even with heavy background clutters can be well located by our P-map powered detection.

\section{Cutout Mask Extraction}
\label{sec:grabcut}

Given a P-map generated by DNCS within a proposal rectangle, our goal is now to generate a binary cutout mask for the object of interest contained therein.
We achieve this goal by adapting the iterative graphcut approach (GrabCut) of Rother et al.~\cite{Rother_SIGGRAPH04_GrabCut}.

The original GrabCut algorithm uses the bounding rectangle to initialize two GMM color models, one for the background, based on colors outside the rectangle, and one for the foreground, based on colors inside the rectangle.
The minimum graphcut is then computed \cite{Kolmogorov:2004}, using the two color models to determine the unary (data) term for each pixel.
The process is then repeated iteratively using the result from the previous iteration to update the background and foreground GMMs, instead of the initial rectangle.

The above process will generally fail to converge to an accurate cutout mask whenever there is a significant overlap between the background and foreground color models, which will happen if the background contains objects with similar colors to those of the foreground object, as demonstrated in Fig. \ref{fig:teaser}(e).
However, armed with our P-map we can initialize the background and foreground color models in a much more precise fashion.

Specifically, we first convert the continuous P-map into an initial binary foreground mask, by computing the minimum graphcut where the unary term at each pixel is determined by our P-map.
Denoting by $p_i$ the P-map value of pixel $i$, we set the foreground likelihood to $P^F_i = p_i^\alpha$ and the background likelihood to $P^B_i = (1-p_i)^\alpha$, where $\alpha = 2.3$.
The resulting binary mask is then used to initialize the two GMM color models, instead of the bounding rectangle.
In subsequent iterations, we set the unary term to a weighted combination of the value predicted by the GMM color model and the P-map likelihood, with the latter's weight decreasing as the iterations progress:
\begin{eqnarray}
  \textit{CP}^F_i & = & \textit{GMM}^F_i \exp(-w P^B_i) \nonumber \\
  \textit{CP}^B_i & = & \textit{GMM}^B_i \exp(-w P^F_i),
\end{eqnarray}
where $\textit{GMM}^F$ and $\textit{GMM}^B$ are the color models for the foreground and background, respectively.
The weight $w = b/k$, where $k$ is the iteration number and $b = 25$ was empirically tuned to reduce the influence of $P^F$ and $P^B$ as the iterations progress.

Fig. \ref{fig:teaser}(d,e) compares two results produced using our P-map enhanced GrabCut (d) with those of the original GrabCut approach (e). It may be seen that the  latter includes in the cutout mask parts of the background which have similar appearance to the foreground chair (in fact, these are parts of identical chairs in the background), while our approach produces a nearly perfect cutout mask.

\section{Experiments}
\label{sec:experiments}

In this section, we quantitatively evaluate the performance of our instance cutout approach, and compare it against several other baseline methods. We also quantitatively evaluate the boost in object detection performance enabled by the use of our P-maps.

\subsection{Evaluation of Instance Cutout}

\begin{table}[t!]
\caption{IoU comparison of various instance cutout probability map generation methods with various post-processing methods. In \emph{ShapeView}, the cutout probability map is generated by rendering images of similar shape retrieval in estimated viewpoint. \emph{DeconvNet (original)} is the DeconvNet model trained on 21 classes of images from PASCAL VOC 2012. \emph{GrabCut + P} is our method described in Section~\ref{sec:grabcut}.}
\label{tab:cutout_mask_comparison}
\begin{center}
\scalebox{1.0}{
\begin{tabular}{ c }
	\includegraphics[width=\linewidth]{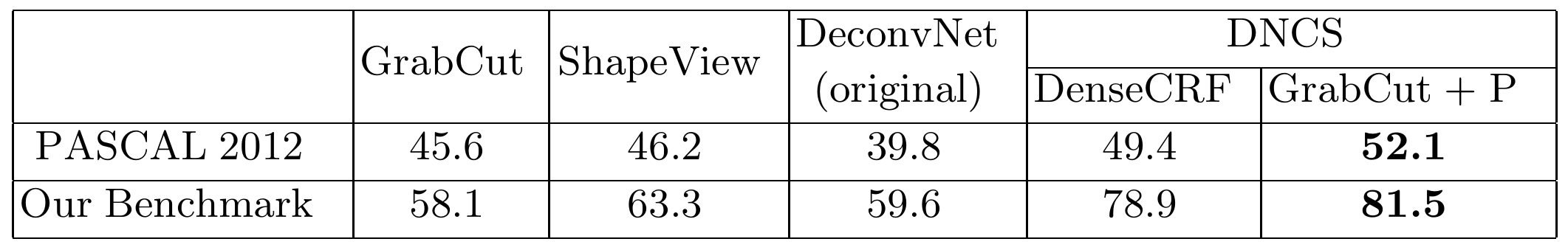} \\
\end{tabular}
}
\end{center}
\end{table}

\paragraph{Dataset and evaluation metric.} We evaluate our instance cutout performance on two chair image datasets. One is from PASCAL VOC 2012, which contains $175$ chair images with ground truth cutout annotations. We found this dataset to be highly challenging for the cutout task, as it contains not only background clutter, but also heavy occlusion, thus many of the chair instances are only partially visible. Occlusion also makes it more challenging for object detection to providing reasonably good proposals, since a rather complete presence of the object of interest is expected. In addition, we have prepared another benchmark, with $418$ chair images, which contains considerable background clutter, but fewer occlusions. We evaluate different approaches using the Intersection over Union (IoU) metric, which measures the ratio between the areas of intersection and union of ground truth and predicted cutout masks. Higher IoU score indicates better cutout accuracy.

\begin{figure}[t!]
\centering
\includegraphics[width=0.8\columnwidth]{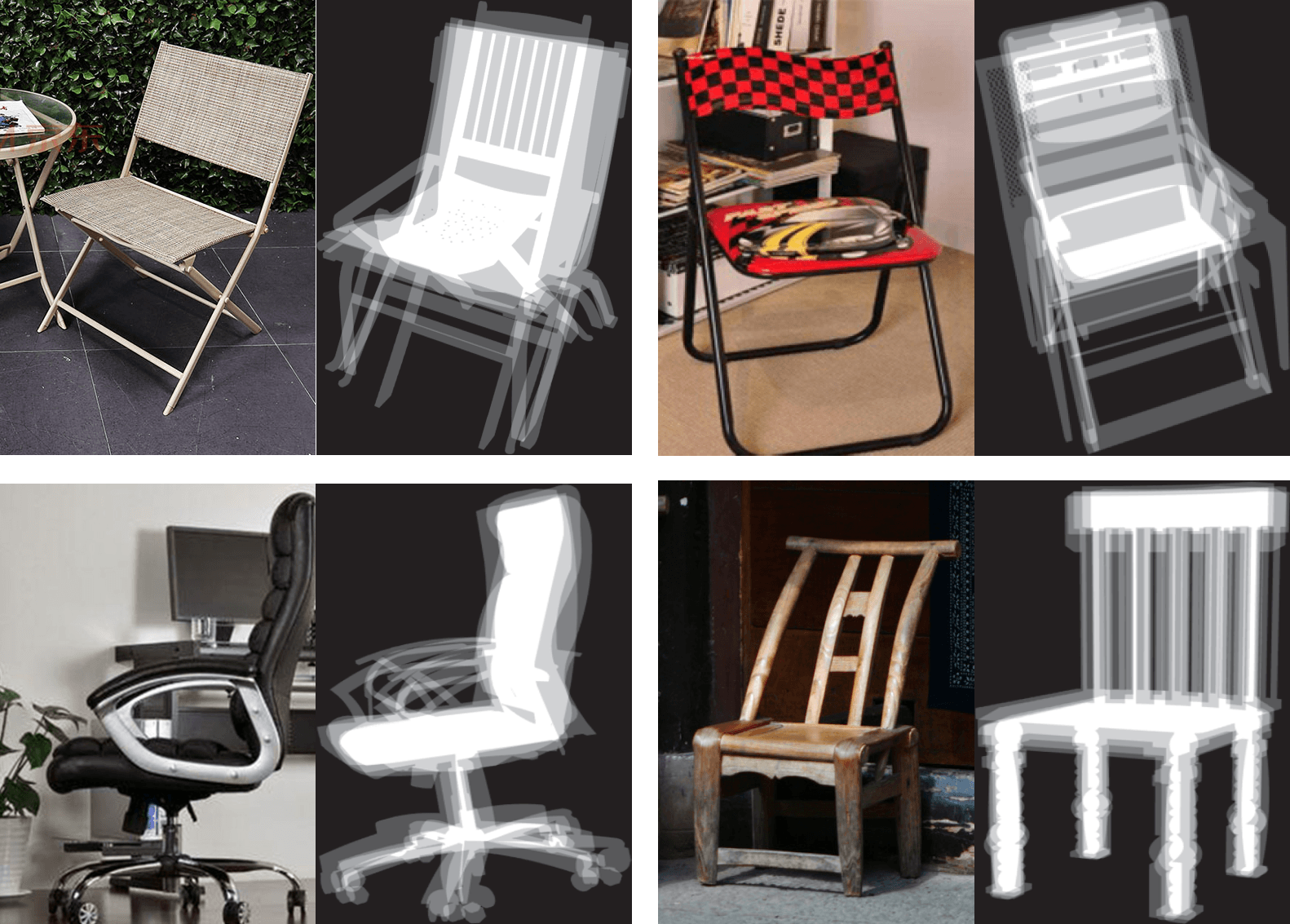}
\caption{Examples of instance probability maps generated by retrieving similar shapes and rendering them from the corresponding predicted viewpoints (ShapeView probability maps).}
\label{fig:shapeview_probmap}
\end{figure}

\paragraph{Baseline methods.} Recent advances in image based 3D object retrieval~\cite{Li_SIGGRAPHAsia15_Joint} and object view estimation~\cite{Su_ICCV15_Render} provide an potential solution for generating an instance probability map, by retrieving similar shapes and rendering them from the predicted viewpoint. The rendered images approximate the underlying object in the input image, and thus can be used as probability maps for instance cutout. More specifically, we pick top $5$ retrievals, render them as binary images from the predicted viewpoints, weight the rendered images by the retrieval confidence, and then overlay them into a normalized instance cutout probability map. We refer to this approach as ``ShapeView''; see Fig.~\ref{fig:shapeview_probmap} for examples of the resulting instance probability maps. Another baseline to our approach are the probability maps generated by the original DeconvNet, which was trained for semantic segmentation with 21 classes. In our comparisons we use GrabCut \cite{Rother_SIGGRAPH04_GrabCut} to generate a cutout mask directly from a given image with a proposal rectangle, while DenseCRF is used for generating a cutout mask from instance probability maps.

We compare our P-map enhanced GrabCut method (Section \ref{sec:grabcut}) applied on the P-maps generated by DNCS model against the original GrabCut, and DenseCRF applied on probability maps generated using the ShapeView approach and the original DeConvNet. The quantitative results are summarized in Table~\ref{tab:cutout_mask_comparison}, while Fig. \ref{fig:comparison} shows a visual comparison using nine examples from our benchmark.
Note that our method outperforms the baseline methods on both the PASCAL VOC 2012 dataset (by $5.9\%$) and on our benchmark (by $18.2\%$) (see Table~\ref{tab:cutout_mask_comparison}). The full set of the test images and the results of these methods is included in our supplementary materials. The performance boost on our benchmark is much higher, since our network was trained with synthetic images that exhibit considerable background clutter, but no occlusions. This suggests an interesting future work direction on synthesizing images with realistic occlusion patterns for training occlusion-aware DNNs. Note that the ShapeView baseline method we proposed also consistently outperforms the original DeConvNet. This may be explained by the fact that it is trained on many classes, and thus cannot learn a sufficiently strong shape prior for each class.

\subsection{Evaluation of Object Proposal Evaluation}

\begin{table}[t!]
\caption{Object proposal evaluation accuracy of classifiers $\mathcal{X}_\textit{SVM}$ and $\mathcal{X}_\textit{CNN}$ on RGB images and RGB-P images. Augmenting the image with a P-channel boosts the performance of both classifiers.}
\label{tab:proposal_evalution_comparison}
\begin{center}
\begin{tabular}{ | c | c | c |}
  \hline
  & RGB Images & RGB-P Images \\ \hline
  $\mathcal{X}_\textit{SVM}$ & 69.6 & 87.9 \\ \hline
  $\mathcal{X}_\textit{CNN}$ & 65.6 & 86.5 \\
	\hline
\end{tabular}
\end{center}
\end{table}

We evaluate the performance of the $\mathcal{X}_\textit{SVM}$ and $\mathcal{X}_\textit{CNN}$ classifiers described in Section~\ref{sec:object_detection} on $35154$ proposals generated by the Selective Search method~\cite{Uijlings_IJCV13_Selective}. These proposals were generated from $52$ images from our benchmark, with each of the images containing a single chair. We measure the accuracy by the average recall on positive and negative samples.

We compare our P-map enhanced $\mathcal{X}_\textit{SVM}$ and $\mathcal{X}_\textit{CNN}$ classifiers against those trained without P-maps, and found that the use of P-maps greatly enhances proposal evaluation accuracy, as reported in Table~\ref{tab:proposal_evalution_comparison}. Our experiment suggests that the instance cutout task should be more tightly coupled with object detection tasks, as the improvement in one benefits the other.

\subsection{Comparison to Seeing 3D Chairs}

\begin{table}[t!]
    \caption{Comparison of top-k detection accuracy between Seeing 3D Chairs, and our P-map powered detection pipeline.}
	\label{tab:seeing_3d_chairs}
	\begin{center}
		\scalebox{1.0}{
			\begin{tabular}{ | c | c | c | c | c | c | }
				\hline
				& Top-1 & Top-2 & Top-3 & Top-4 & Top-5 \\
				\hline
				Seeing 3D Chairs & 13.86 & 24.67 & 28.11 & 28.78 & 30.61 \\
				\hline
				Selective Search + $\mathcal{X}_\textit{SVM}$ & 21.73 & 28.76 & 35.49 & 40.16 & 43.49 \\
				\hline
				Selective Search + $\mathcal{X}_\textit{CNN}$ & 20.29 & 31.58 & 38.41 & 44.86 & 49.37 \\
				\hline
			\end{tabular}
		}
	\end{center}
\end{table}

We also compare chair detection performance based on Selective Search + $\mathcal{X}$ with that proposed in Seeing 3D Chairs~\cite{Aubry_CVPR14_Seeing}. Given an image, Seeing 3D Chairs outputs a ranked list of chair proposals. We generate chair proposals with Selective Search and then rank them with our classifiers. We compare the top-k detection accuracy of these approaches on the first $100$ chair images from PASCAL VOC 2012. The results are reported in Table~\ref{tab:seeing_3d_chairs}. Note that Seeing 3D Chairs is also an approach extensively trained on the chair class, yet we show that our P-map powered approach achieves better accuracy.

\begin{figure}[t!]
\centering
\includegraphics[width=\linewidth]{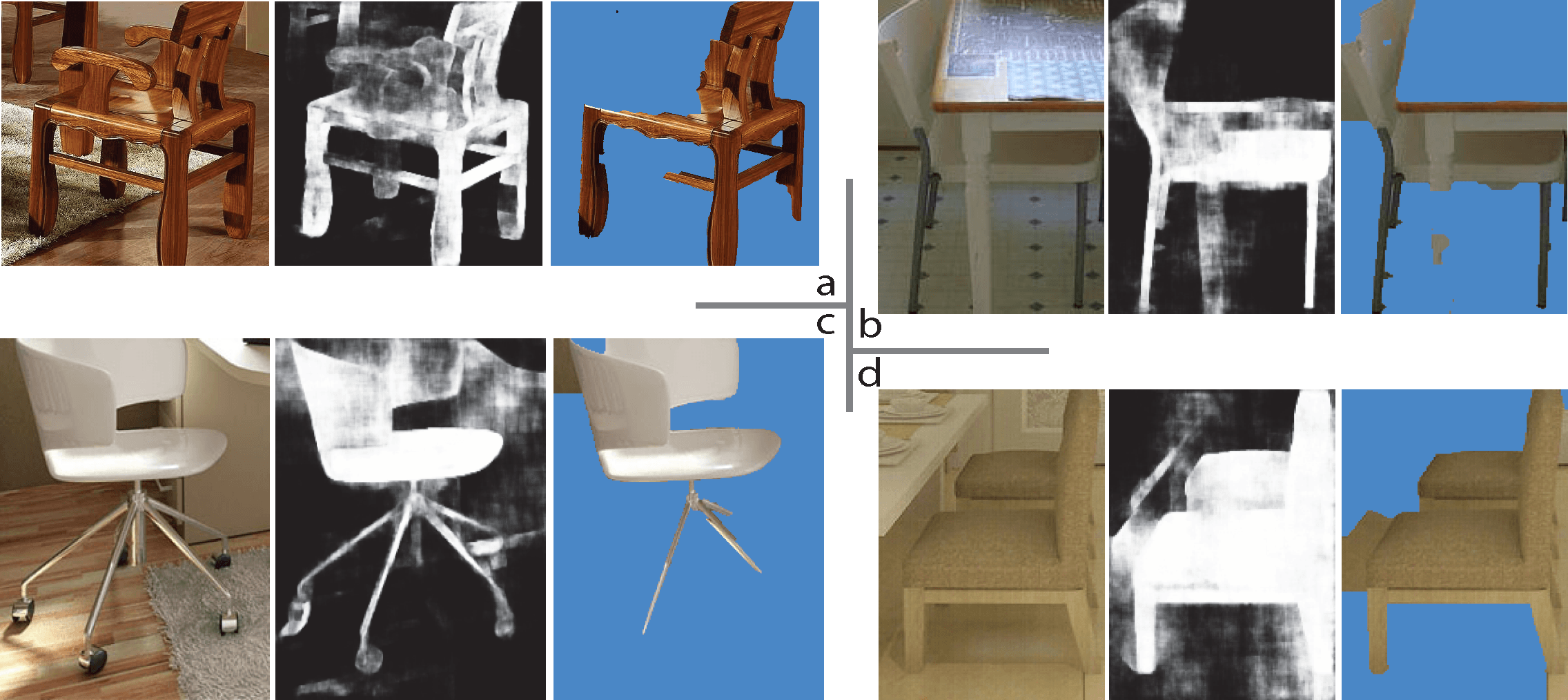}
\caption{Failure cases. We found several sources of errors in our cutout masks: (a) Chairs that are rarely seen in training data might be misunderstood by the DNN; (b) Occlusions pose additional challenges over background clutter; (c) The binary mask generation step sometimes eliminates thin structures even though they are preserved in the probability map; (d) Strong similarities between objects might result in highly confusing situation from specific view points.}
\label{fig:failure_cases}
\end{figure}

\section{Conclusions}
\label{sec:conclusions}

Many computer graphics applications depend on accurate object cutouts. Facilitating automatic cutout extraction remains extremely challenging, since it cannot rely on low-level image analysis alone, and necessarily requires some degree of high-level semantic analysis. The P-maps that we introduced aim to provide some of the latent semantics to assist in the extraction of cutouts. The presented network aims to encode in the P-maps the essence of the shape prior with rich variability of geometry and topology.

The semantic information that P-maps carry was shown to be effective not only directly for cutouts, but also for locating the target object. We have shown that they significantly improve the evaluation of proposals, which are again means to enhance and accelerate a variety of applications that require image analysis.

The claim to fame of the P-maps is their competence to deal with cluttered images, where the target object has ``rivals'' in its background. Our network was designed explicitly to deal with these types of distractions, and together with our modified GrabCut approach makes a substantial step toward automatic and accurate instance cutout.

Nevertheless, our approach has its limitations. First, it is category specific, and requires training on the target class. It is intensively data-driven, which implies that a large amount of annotated data is required. For chairs, the problem is less significant since large 3D datasets are readily available. However, there are always peculiar shapes (see Fig.~\ref{fig:failure_cases} (a)). For many other object classes there is no comparable availability of rich enough 3D models, yet. Second, the relative size of target object in the input image should be in an expected range, defined by the training set. Arguably, a more significant limitation of our technique is occlusion (see Fig.~\ref{fig:failure_cases} (b)). While cluttering is handled well, occlusion remains a hurdle. For this reason, our performance advantage on the challenging PASCAL VOC 2012 benchmark is somewhat more modest.
One of the challenges we encountered in training for occlusion is to realistically synthesize it, which is left for future work. Another limitation is demonstrated in Fig.~\ref{fig:failure_cases} (c), where the final binary mask generation step sometimes fails capture thin structures, even though they are present in the P-map.

We believe that more fundamental processing can benefit from similar semantic layers. For example, image-based 3D shape retrieval, 2D-3D correspondence, or fitting and registering 3D proxies into an image. The P-maps or possibly similar semantic layers have the potential to boost the performance of applications that link 2D to 3D. We would also like to explore the potential of P-maps for enhance other low-level image processing operations, such as edge detection, where the saliency of the edge is augmented or amplified by the P-channel.


\vspace{3mm}
\noindent {\bf Acknowledgement} This work was supported by National 973 Program(2015CB352501), NSFC-ISF(61561146397), Shenzhen Knowledge innovation program for basic research (JCYJ20150402105524053).

\bibliographystyle{splncs}



\end{document}